%% file: main.tex
\documentclass[10pt,twocolumn,letterpaper]{article}
\usepackage{iccv}              

\usepackage{bm} 
\usepackage{array}
\usepackage{adjustbox}
\usepackage{multirow, makecell}
\usepackage{pifont}
\usepackage{comment}

\newcolumntype{C}[1]{>{\centering\arraybackslash}p{#1}}


\definecolor{iccvblue}{rgb}{0.21,0.49,0.74}
\usepackage[pagebackref,breaklinks,colorlinks,allcolors=iccvblue]{hyperref}

\def\paperID{*****} 
\def\confName{ICCV}
\def\confYear{2025}

\title{LOTA: Bit-Planes Guided AI-Generated Image Detection}

\author{
	Hongsong Wang\textsuperscript{1,2 * \textdagger}, Renxi Cheng\textsuperscript{3 \textdagger}, Yang Zhang$^{4}$, Chaolei Han$^{3}$, Jie Gui\textsuperscript{3,5,6}\thanks{Corresponding authors. \textsuperscript{\textdagger}{Equal Contribution}}\\
	$^{1}$School of Computer Science and Engineering, Southeast University, Nanjing 210096, China \\
	$^2$Key Laboratory of New Generation Artificial Intelligence Technology and Its Interdisciplinary \\
	Applications (Southeast University), Ministry of Education, China \\ 
	$^{3}$School of Cyber Science and Engineering, Southeast University, Nanjing 210096, China\\
	$^{4}$School of Computer Science and Software Engineering, Shenzhen University, Shenzhen 518060, China \\
	$^{5}$Purple Mountain Laboratories, Nanjing 210000, China \\
	$^{6}$Engineering Research Center of Blockchain Application, Supervision And Management\\ (Southeast University), Ministry of Education, China \\
	\tt\small\{hongsongwang, renxi, chaoleihan, guijie\}@seu.edu.cn, yangzhang@szu.edu.cn \\
}

\begin{document}
\maketitle
\input{sec/0_abstract}    
\input{sec/1_introduction}

\input{sec/2_related_work}
\input{sec/3_method}
\input{sec/4_experiments}
\input{sec/5_conclusion}  

\section*{Acknowledgments}
This work was supported by National Science Foundation of China (62172090, 62302093, 62176163), Jiangsu Province Natural Science Fund (BK20230833), Start-up Research Fund of Southeast University under Grant RF1028623097, Shenzhen Higher Education Stable Support Program General Project (Grant 20231120175215001), Scientific Foundation for Youth Scholars of Shenzhen University, and Big Data Computing Center of Southeast University.

{
    \small
    \bibliographystyle{ieeenat_fullname}
    \bibliography{main}
}

\end{document}

%% file: sec/0_abstract.tex
\begin{abstract}

The rapid advancement of GAN and Diffusion models makes it more difficult to distinguish AI-generated images from real ones. Recent studies often use image-based reconstruction errors as an important feature for determining whether an image is AI-generated. However, these approaches typically incur high computational costs and also fail to capture intrinsic noisy features present in the raw images. To solve these problems, we innovatively refine error extraction by using bit-plane-based image processing, as lower bit planes indeed represent noise patterns in images. We introduce an effective bit-planes guided noisy image generation and exploit various image normalization strategies, including scaling and thresholding. Then, to amplify the noise signal for easier AI-generated image detection, we design a maximum gradient patch selection that applies multi-directional gradients to compute the noise score and selects the region with the highest score. Finally, we propose a lightweight and effective classification head and explore two different structures: noise-based classifier and noise-guided classifier. Extensive experiments on the GenImage benchmark demonstrate the outstanding performance of our method, which achieves an average accuracy of \textbf{98.9\%} (\textbf{11.9}\%~$\uparrow$) and shows excellent cross-generator generalization capability. Particularly, our method achieves an accuracy of over 98.2\% from GAN to Diffusion and over 99.2\% from Diffusion to GAN. Moreover, it performs error extraction at the millisecond level, nearly a hundred times faster than existing methods. The code is at https://github.com/hongsong-wang/LOTA.
\end{abstract}


%% file: sec/1_introduction.tex
\section{Introduction}
\label{sec:introduction}

With the rapid development of generative models, especially Generative Adversarial Networks (GANs) \cite{goodfellow2014generative} and Diffusion models \cite{rombach2022high}, AI-generated images are becoming more and more realistic, and it is even difficult for people to distinguish the difference between real and AI-generated images. These AI-generated images may also be used for illegal purposes \cite{juefei2022countering,cao2025external}, such as spreading unreal information or harmful content, which may mislead or harm the public. Therefore, there is an urgent need for a robust technique for distinguishing AI-generated images from real ones.

\begin{figure}[t]
  \centering
  \small
  \includegraphics[width=\linewidth]{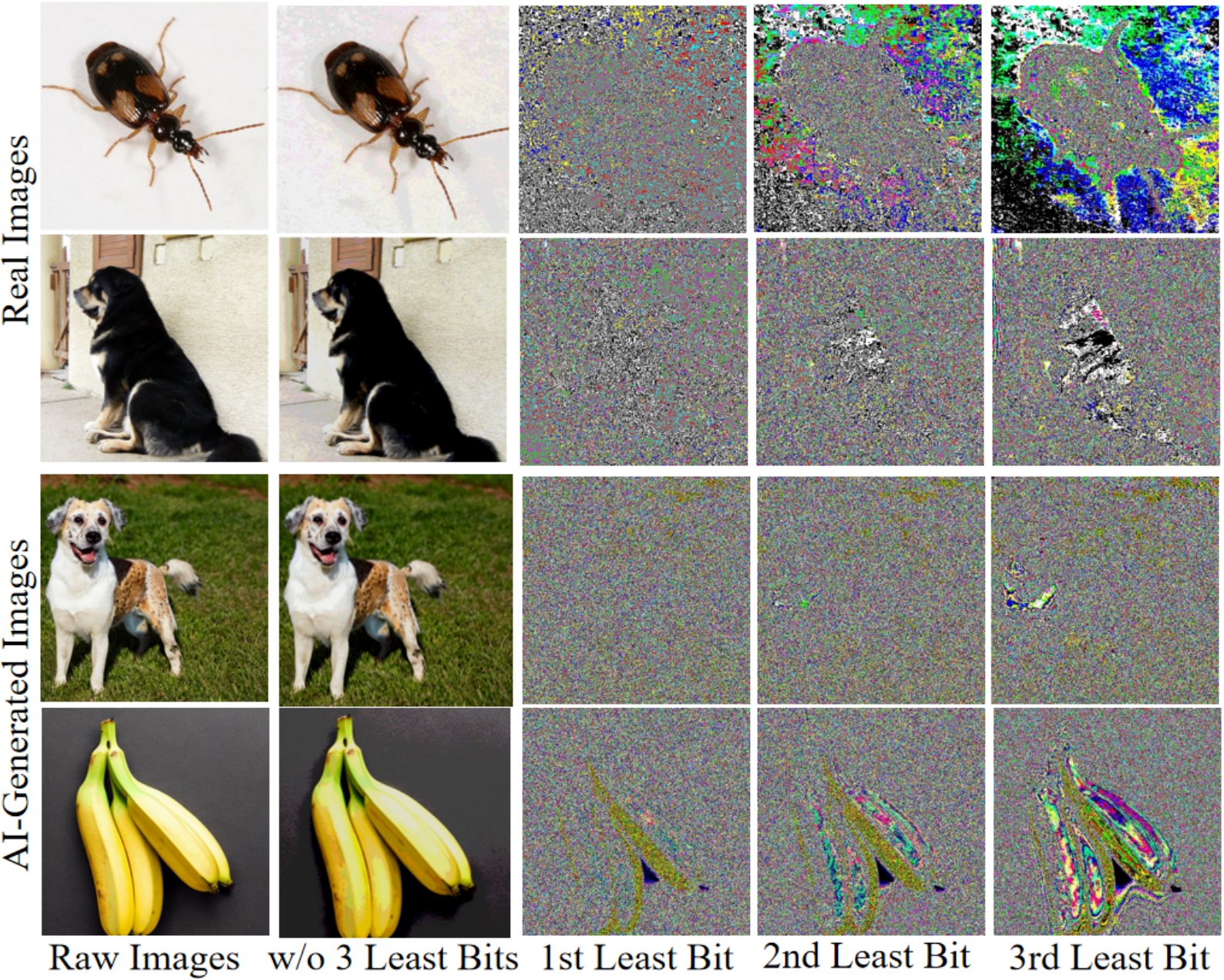}
  \caption{\textbf{Comparison of least bit-planes between real images and AI-generated images.} We extract the 1st, 2nd and 3rd least bit-planes from both types of images, separately. We find that images of low bit-planes of real images and AI-generated images have different noise patterns and distributions that can be used for distinguishing between them. Low bit-planes of fake images contain artifacts that are invisible in RGB images.
  }
  \label{fig:motivation}
\end{figure}

Early deep learning-based deepfake methods focus on the detection of GAN-generated images. However, recent works \cite{corvi2023detection,ricker2022towards} find that with the emergence of diffusion models, detection performance declines significantly when GAN-based detection methods are applied to diffusion-generated images. With regard to detecting diffusion-generated images, many works are based on reconstructed image errors, \textit{e.g}., DIRE \cite{wang2023dire}, SeDID \cite{ma2023exposing}, LaRE$^2$ \cite{luo2024lare}, ESSP \cite{chen2024single}, ZED \cite{cozzolino2024zero}. For example, DIRE \cite{wang2023dire} computes the error between the raw and reconstructed images, and considers the image with smaller error to be AI-generated. SeDID \cite{ma2023exposing} computes the loss error for a given step in the forward and reverse process of Diffusion, and regards the image with smaller error as AI-generated. These approaches extract error signals through a multi-step DDIM sampling process, which is not only inefficient but also susceptible to introducing random noise at different steps. 

Least Significant Bit (LSB)-based steganography is a simple yet effective technique that embeds a secret message in pixel values while minimizing perceptible distortions~\cite{johnson1998exploring}. LSB-based steganography can be easily extended to multiple bit-planes, with lower ones prioritized to preserve perceptual quality and local properties checked when using higher bit-planes, while the \textit{k}-least significant bits can also be utilized for steganography~\cite{kawaguchi1999principles, elharrouss2020image}. However, most LSB-based methods are primarily limited to the fields of image steganography and steganalysis. 

Bit-planes based approaches have the potential to address AI-generated image detection, as they are capable of exploiting subtle differences in pixel values and detecting artifacts that are typically absent in natural images. We visualize and compare the least bit planes for both real and AI-generated RGB images in Figure~\ref{fig:motivation}. It can be seen that, although removing the least significant three bit-planes has almost no impact on the visual appearance of both real and AI-generated RGB images, differences still exist in the bit-plane images between real and AI-generated RGB images. Compared to real images, the brightness of the least significant bit-planes in AI-generated images is low, or the brightness distribution is irregular. Moreover, low bit-planes of fake images contain artifacts. 
One possible reason is that current image generation models lack the capability to generate visually imperceptible details. Since there are no such works on AI-generated image detection, we aim to fill this gap and leverage imperceptible bit-planes for this purpose.

To this end, we introduce a simple yet effective approach called LOw-biT pAtch (LOTA) for detecting AI-generated images. LOTA consists of three key modules: Bit-planes Guided Noisy Image Generation (BGNIG), Maximum Gradient Patch Selection (MGPS) and classification head. BGNIG takes lower bit-planes, which contains noise, to extract the error image and achieves high efficiency and accuracy. To enhance the brightness of the noisy image, we explore two normalization methods: scaling and thresholding. To further amplify the noise signal for subsequent detection, the MGPS calculates the noise score using multi-directional gradients and selects the patch with the highest score. Finally, we introduce Noise-Based Classifier (NBC) and Noise-Guided Classifier (NGC). The NBC is a simple convolutional neural network based solely on the noise image, while the NGC uses noise patches to effectively guide fake detection from the raw image.
We conduct extensive experiments on GenImage \cite{zhu2023genimage}, where images are generated by eight different generators. Compared to existing mainstream methods, our approach is more effective, faster, and more generalizable. The main contributions are as follows:

\begin{itemize}
\item[$\bullet$] \textbf{Novel solution for AI-generated image detection:} We innovatively address AI-generated image detection based on bit-planes, and propose an efficient approach for noisy representation extraction. 
\item[$\bullet$] \textbf{Efficient pipeline design:} We propose a simple yet effective pipeline with three modules: noise generation, patch selection and classification. We design a heuristic strategy called maximum gradient patch selection and introduce two effective classifiers: noise-based classifier and noise-guided classifier. Our approach operates at millisecond level, nearly a hundred times faster than current methods.
\item[$\bullet$] \textbf{Exceedingly superior performance:} Extensive experiments demonstrate the effectiveness of LOTA, which achieves \textbf{98.9\%} ACC on GenImage, showing great cross-generator generalization capability and outperforming existing mainstream methods by more than \textbf{11.9\%}.
\end{itemize}

%% file: sec/2_related_work.tex
\section{Related Work}
\label{sec:related_work}
Recent advances in AI-generated image detection have focused on exploiting artifacts across different domains. We briefly review works belonging to the following categories.

\noindent\textbf{Spatial Domain-Based Methods:} Most detecting methods are based on the spatial domain. Early studies primarily analyze pixel-level texture patterns and geometric inconsistencies. Wang \textit{et al.} \cite{wang2020cnn} demonstrate that CNN-generated images tend to exhibit distinguishable artifacts that can be detected. Subsequent studies extend this to real-image priors \cite{LiuLNP} and generalized gradient artifacts \cite{tan2023learning}. With the rise of Diffusion models \cite{rombach2022high}, DIRE \cite{wang2023dire} considers the reconstruction error as an essential metric for detecting generated images, and GLFF \cite{ju2023glff} fuses global and local features to capture multi-scale inconsistencies. 
Recently, DRCT \cite{chen2024drct} realizes universal detection through contrastive reconstruction. ESSP \cite{chen2024single} attains outstanding performance by using single-patch analysis. Additionally, geometric inconsistencies \cite{sarkar2024shadows} and zero-shot frameworks \cite{cozzolino2024zero} further extend spatial domain analysis.

\noindent\textbf{Frequency Domain-Based Methods:} Frequency analysis reveals artifacts often imperceptible in pixel space. The seminal work of \cite{zhang2019detectingSPEC} identified GAN-specific frequency artifacts. Later, Dzanic \textit{et al.} \cite{dzanic2020fourier} focus on high-frequency features to better simulate real images. Chandrasegaran \textit{et al.} \cite{chandrasegaran2021closer} validate these findings for existing CNN-based generative models. Corvi \textit{et al.} \cite{corvi2023detection, corvi2023intriguing} extend frequency analysis to Diffusion models by exploring distinct fingerprints and differences. Recent frequency masking techniques \cite{doloriel2024frequency} further enhance generalization across different generators. These methods face challenges in handling high-resolution images and adaptive generation strategies.

\noindent\textbf{Multi-Domain Feature Fusion-Based Methods:} Traditional methods have difficulties coping with increasingly high-quality generated images, so several studies try to integrate complementary signals from multiple domains. Yu \textit{et al.} \cite{yu2022detection} combine texture and semantic features to detect manipulated faces, and Luo \textit{et al.} \cite{luo2024lare} combine reconstructed error features and latent space features to detect generated images. Efficiency is also prioritized. Lanzino \textit{et al.} \cite{lanzino2024faster} employ binary neural network by combining frequency-domain features, local texture features and pixel-domain features. Leporoni \textit{et al.} \cite{leporoni2024guided} fuse RGB and depth features to exploit 3-Dimension inconsistencies. These methods demonstrate that multi-domain fusion enhances robustness against evolving generation techniques.

\noindent\textbf{Image Error-Based Methods:}
Regarding the detection of diffusion-generated images, several recent methods rely on error computation, focusing on the differences between reconstructed and raw images. DIRE \cite{wang2023dire} calculates reconstructed image errors to distinguish generated images from real ones. SeDID \cite{ma2023exposing} extracts the loss error for a given processing step in the Diffusion process. LaRE$^2$ \cite{luo2024lare} first computes noise image within the diffusion-based framework, then introduces both spatial and channel feature refinement to enhance feature learning. Chen et. al \cite{chen2024single} exploit the noise pattern of an image for detection and confirm that more noise indicates a higher possibility of being real images. Cozzolino et. al \cite{cozzolino2024zero} calculate the differences between expected and actual coding cost of an image for detection. Different from these approaches, we attempt to extract the noise signal contained within the image itself. 

%% file: sec/3_method.tex
\begin{figure}[t]
  \centering
  \small
  \includegraphics[width=1\linewidth]{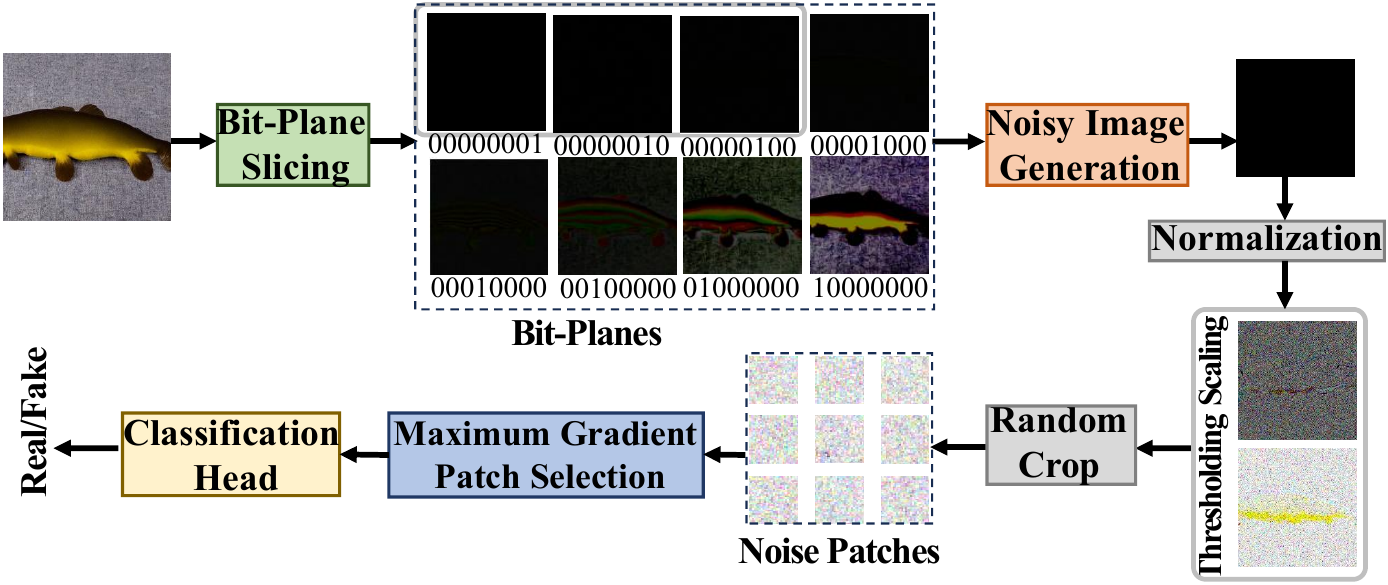}
  \caption{\textbf{Overview of our method.} First, we decompose the image into 8 bit-planes, and compose least bit-planes to generate the noise representation. Second, we crop the noise image into several patches, and select the patch with the highest gradient-based score. Finally, a classification head is applied.}
  \label{fig:overview}
\end{figure}

\section{Method}
We introduce LOTA for AI-generated image detection. LOTA comprises three subsequent modules: Bit-Planes Guided Noisy Image Generation, Maximum Gradient Patch Selection and classification head. An illustration of LOTA is provided in Figure~\ref{fig:overview}, with details described below.

\subsection{Bit-Planes Guided Noisy Image Generation}
The reconstruction error images, such as those from DIRE \cite{wang2023dire}, LaRE$^2$ \cite{luo2024lare} and ESSP \cite{chen2024single}, extract error maps or  noise patterns,that can be effectively utilized for detecting generated images. Inspired by this, we use noise images for generated image detection. However, instead of relying on reconstruction error, we exploit the noise map inherently contained in the image. 

\begin{figure}[t]
  \centering
  \small
  \includegraphics[width=1\linewidth]{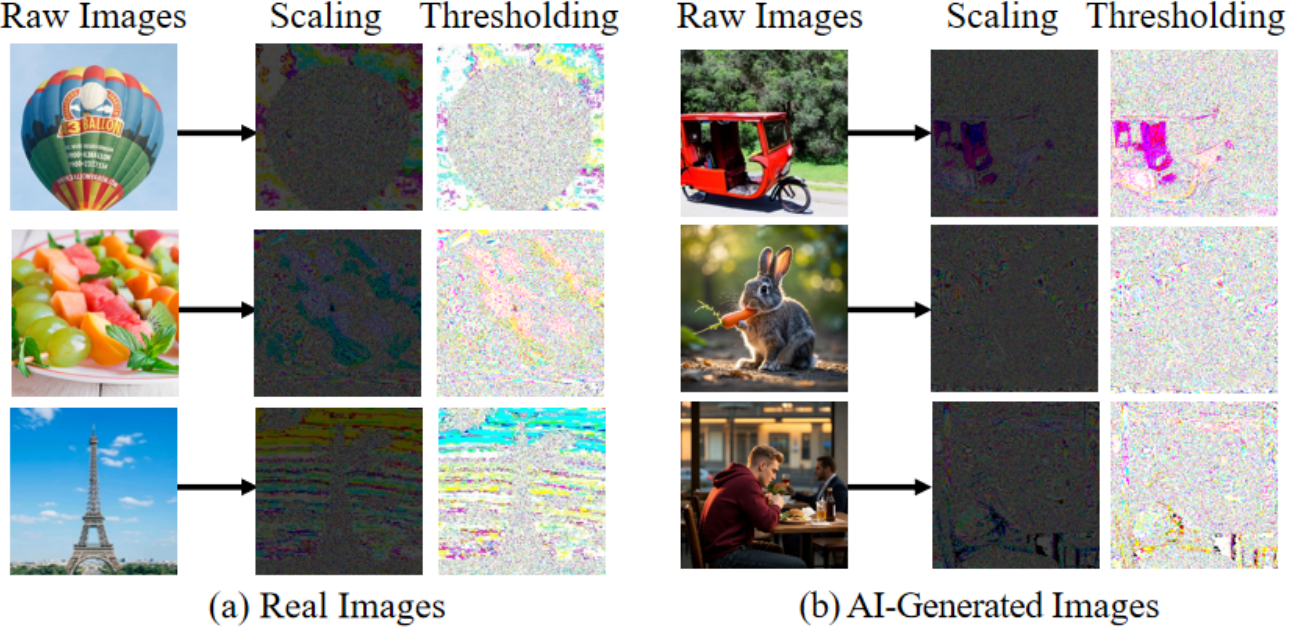}
  \caption{\textbf{Visualizations of generated noisy images by scaling and thresholding in the BGNIG module.} We compare corresponding noise images for real and AI-generated images. Both scaling and thresholding methods effectively extract the noise patterns of images. The brightness distribution of noisy images from real images is relatively regular. However, noisy images from synthetic images contain several regions with artifacts.
  }
  \label{fig:visualization}
\end{figure}

A bit-plane of an image consists of the bits at a specific position in the binary representation of each pixel. Since a gray-scale image is typically represented with eight bits per pixel, it contains eight bit-planes in total. Then a RGB image can be seem to be composed of three gray-scale images of each channel. Let $\bm{x^c}$ be the RGB image of the channel $c$, where $c \in \left\{ {R,G,B} \right\}$, and let $\bm{x_k^{c}}$ denote its corresponding $k$-th bit-plane of the channel $c$, where $0 \leq k \leq 7$, then the image of this channel can be decomposed as:
\begin{equation}
  \bm{x^{c}} = \sum\limits_{k = 0}^7 {2^k} \cdot \bm{x_k^{c}}.
  \label{eq:decomposition}
\end{equation}
Dividing an image into multiple bit planes is known as bit-plane slicing. Higher-order bit-planes contain visual information such as textures and colors, while lower-order bit-planes preserve details including contours and noises.

To extract noise patterns in images, we select the three lowest-order bit-planes of each channel, namely $\bm{x_2^{c}}$, $\bm{x_1^{c}}$, and $\bm{x_0^{c}}$, to generate a low-bit image. The lowest-order bit-planes of the image are composed with addition operations. Specifically, the following formulation is applied:
\begin{equation}
  \bm{z^{c}} = {2^2} \cdot \bm{x_2^{c}} + {2} \cdot \bm{x_1^{c}} + \bm{x_0^{c}},
  \label{eq:composition}
\end{equation}
where $\bm{z^{c}}$ denotes the composed low-bit image for each channel of the RGB image.

Since the pixel values in $\bm{z^{c}}$ range from 0 to 7, normalization needs to be applied before extracting image features. Two distinct methodologies are employed: scaling and thresholding.

\noindent\textbf{Scaling: } 
The min-max normalization is used to scale these values to [0, 255]:
\begin{equation}
  \bm{\tilde{z}^{c}} = 255 \cdot \frac{{\bm{z^{c}} - \bm{z_{\min}^{c}}}}{\bm{{z_{\max}^{c}} - \bm{z_{\min}^{c}}}},
\end{equation}
where $\bm{\tilde{z}^{c}}$ denotes the $c$-th channel of normalized noise $\bm{\tilde{z}}$.

\noindent\textbf{Thresholding: }
Since the values in the lower bit planes are sparse, we mitigate this issue by directly setting all values greater than 0 to 255. This approach is called thresholding, which enhances the brightness of the normalized image. The formulation of thresholding is:
\begin{equation}
  {\tilde{z}}_{i,j}^{c} = \left\{ \begin{array}{l}
0,\quad\,\, \text{if} \,\, z_{{i,j}}^{c} = 0,\\
255,\,\, \text{if} \,\, z_{{i,j}}^{c} > 0,
\end{array} \right.
\end{equation}
where $z_{{i,j}}^{c}$ represents the element at the $i$-th row and $j$-th column of $z^c$ in Eq.~(\ref{eq:composition}).

Figure~\ref{fig:visualization} shows visualizations of noisy images generated by two different approaches for both real and AI-generated images. We observe that for real images, the brightness distribution of noisy images is relatively regular, with visible object contours and some texture information. In contrast, for AI-generated images, the brightness distribution of noisy images is relatively chaotic, making it difficult to discern the contours of objects in original images.


\subsection{Maximum Gradient Patch Selection}
Though the noise pattern in low-bit images serves as a critical feature for distinguishing real and generated images, it still contains much useless information that may interfere with detection. To further extract the intrinsic feature which can distinguish real and generated images in essence, we introduce Maximum Gradient Patch Selection (MGPS) to select the most informative patch from an image for further detection.

For the low-bit images $\tilde{\bm{z}}$, we randomly divide them into non-overlapping patches. Then, we design a divergence-based score function to measure the sparsity of image gradients in different directions. Let $\tilde{z}_p$ be the noisy patch, where $p$ denotes the index of the patch numbers, the score $g_p$ is computed as:
\begin{equation}
\begin{aligned}
    g_p &= \|\bm{\tilde{z}_p} * \bm{g_x}\|_1 + \|\bm{\tilde{z}_p} * \bm{g_y}\|_1  \\
        &+ \|\bm{\tilde{z}_p} * \bm{g_{xy}}\|_1 + \|\bm{\tilde{z}_p} * \bm{g_{yx}}\|_1,
\end{aligned}
\end{equation}
where $*$ represents the image convolution operation, $\|\cdot\|_1$ denotes $L1$ norm of the matrix, $g_x, g_y, g_{xy}$ and $g_{yx}$ are convolution kernels described as:
\[
\begin{aligned}
\bm{g_x} &= \begin{bmatrix} -1 & 1 \end{bmatrix}, \quad\,
\bm{g_y} &= \bm{g_x}^T, \\
\bm{g_{xy}} &= \begin{bmatrix} -1 & 0 \\ 0 & 1 \end{bmatrix}, \quad
\bm{g_{yx}} &= \begin{bmatrix} 0 & -1 \\ 1 & 0 \end{bmatrix}.
\end{aligned}
\]

The first two terms of the score represent horizontal and vertical gradients, while the latter two terms represent the diagonal gradients. Since the patch denotes image noise, the scores with large values often correspond to regions with excessive high-frequency variations, which are likely dominated by noise or structural details rather than image content. For AI-generated images, these high-divergence regions might indicate artifacts caused by imperfect generative models. 

Thus, we select the noise patch with the highest $g_p$ score:
\begin{equation}
\bm{\tilde{z}_{p^*}} = \arg\max_{p} g_p,
\end{equation}
where $p^*$ denotes the index of the best patch.

It should be noted that although both our approach and ESSP \cite{PatchCraft} select a simple patch, our MGPS differs from ESSP in three key aspects. First, we computes the gradient-based score instead of the texture diversity score. Second, we formulate the score function using a concise and efficient image convolution operation. Third, we choose the patch with the highest score instead of the lowest one. 


\begin{figure}[t]
  \centering
  \small
  \includegraphics[width=1\linewidth]{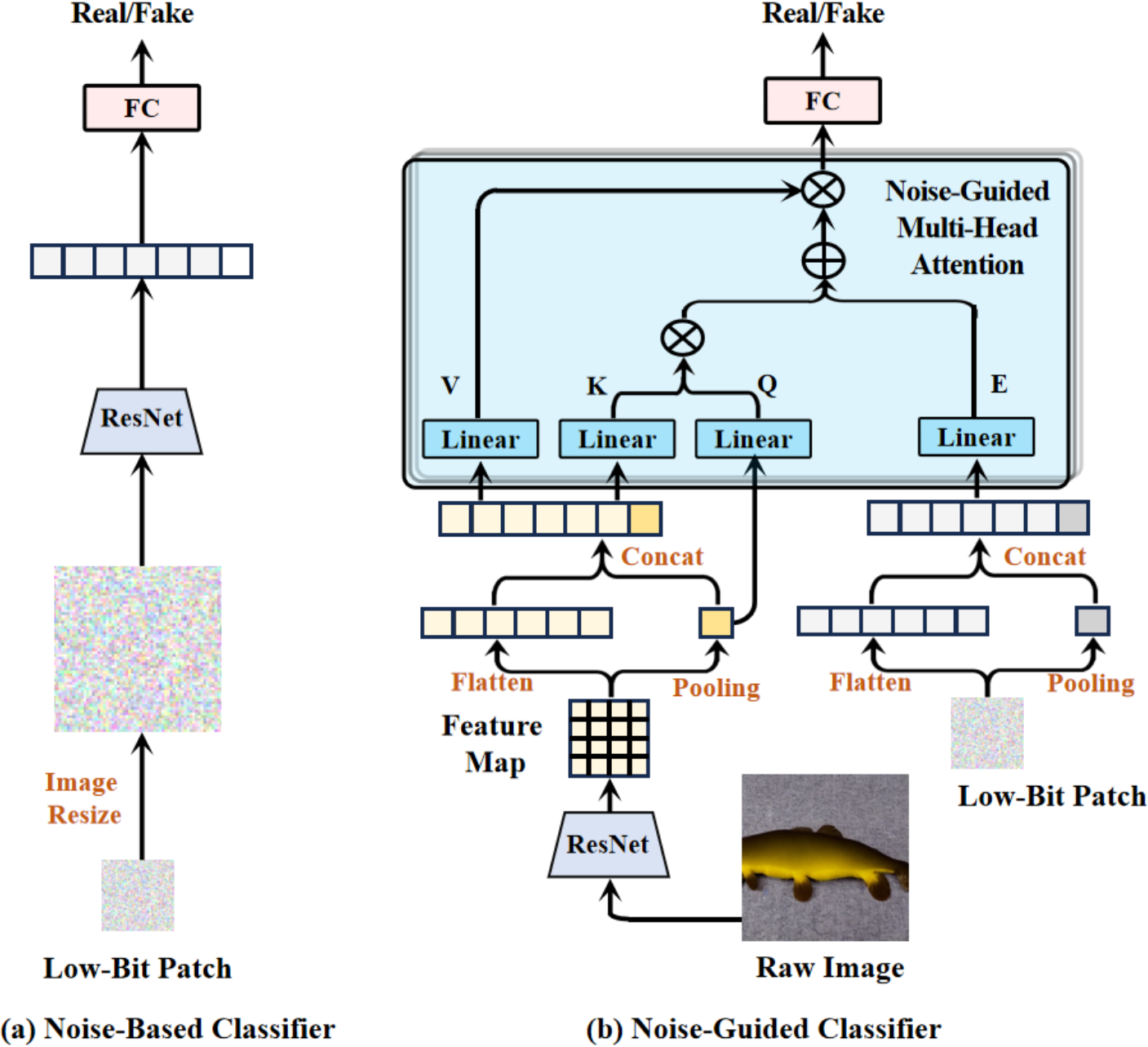}
  \caption{\textbf{Structure of the classifier.} Two different classifiers are applied. For the noise-based classifier, the low-bit patch is directly fed into the ResNet. For the noise-guided classifier, the low-bit patch and the feature map of raw images are combined by using the noise-guided multi-head attention.}
  \label{fig:classifier}
\end{figure}

\subsection{Classification Head}
After obtaining the selected patch of noise image, we present two methods as classifiers for AI-generated image detection. The structures of the two classifiers are illustrated in Figure~\ref{fig:classifier}, with details described below. 

\noindent\textbf{Noise-Based Classifier:} A straightforward approach is to use a convolutional neural network pre-trained on ImageNet as the classifier for low-bit patch images. Since the patch is small, it needs to be resized to the standard size of $256 \times 256$ before being fed into the convolutional classifier. 

\noindent\textbf{Noise-Guided Classifier:} Current studies mainly focus on finding a more effective error extraction method \cite{wang2023dire,ma2023exposing,chen2024single,cozzolino2024zero}, but neglect the information of raw images. So we align the raw and error maps from the spatial perspective to provide more reliable information for classification.

For the raw image $\bm{x}$, we first put it through the image encoder (e.g., ResNet-50) to get the feature map $\bm{\tilde{x}}$, then the pooling, flatten and flatten are respectively used to get query $Q$, key $K$ and value $V$ for the spatial attention. For the error patch $\bm{\tilde{z}_{p^*}}$, it is flatten and projected to get error $E$ for the spatial attention. So the Noised-Guided Multi-Head Attention can be defined as:
\begin{equation}
U = \mathrm{softmax}\left( {\frac{{Q{K^T}}}{{\sqrt {{d_k}} }} + E} \right)V,
\end{equation}
where $\rm{softmax} \left( \cdot \right)$ is the activation function, ${d_k}$ is the dimension of the tensor $K$. Based on this, we further apply the multi-head Attention according to Transformer \cite{vaswani2017attention}. 

The output vector is then followed by a fully connected layer with binary cross-entropy loss to distinguish between real and generated images.

%% file: sec/4_experiments.tex
\section{Experiments}
\subsection{Dataset and Implementation Details}

\noindent\textbf{Dataset and Evaluation Metrics:} We evaluate the proposed method on GenImage dataset \cite{zhu2023genimage}, which employs ImageNet dataset as real images, and incorporates eight mainstream GAN and Diffusion generators (including BigGAN \cite{BigGAN}, Midjourney \cite{Midjourney}, Wukong \cite{wukong}, Stable Diffusion V1.4 \cite{rombach2022high}, Stable Diffusion V1.5 \cite{rombach2022high}, ADM \cite{dhariwal2021diffusion}, GLIDE \cite{nichol2021glide}, and VQDM \cite{gu2022vectorvqdm}) to generate AI-generated images. The dataset comprises a total of 1,331,167 real images and 1,350,000 generated images. The data corresponding to each generator are split into training and testing subsets. For each classifier, training was conducted on the training subsets, followed by comprehensive evaluation across all eight testing subsets. Following existing works for AI-generated image detection \cite{wang2023dire,luo2024lare,chen2024single}, we adopt accuracy (ACC) and average precision (AP) as evaluation metrics. The threshold for computing accuracy is 0.5.

\noindent\textbf{Implementation Details:} Before Maximum Gradient Patch Selection (MGPS), the noise image is resized to a resolution of $256\times256$. In MGPS, a $32\times32$ patch is selected. For noise-guided classifier, the patch is directly fed into the classifier, combined with corresponding raw images. For Noise-Based Classifier, the patch is first resized to $256\times256$. ResNet-50 is used as the image encoder for the classifier. During training, the learning rate is $0.0001$, and the batch size is 64. The maximum number of training epochs is 30, with Adam as the optimizer.


\subsection{Experimental Results}

\begin{table*}[h]
    \centering
    \caption{\textbf{Detailed results on the GenImage dataset.} The model is trained on eight subsets of GenImage and tested on the corresponding subsets. The default LOTA employs thresholding during noisy image generation and Noise-Based Classifier for the classification head. The two other variants, LOTA-\textit{scl.} and LOTA-\textit{ngc}, use scaling and noise-guided classifier, respectively.
    }
    \vspace{-0.5em}
    \label{tab:experiments}
    \adjustbox{max width=0.92\textwidth}{
    \begin{tabular}{lC{1.8cm}C{1.4cm}C{1.4cm}C{1.4cm}C{1.4cm}C{1.4cm}C{1.4cm}C{1.4cm}C{1.4cm}C{1.4cm}}
        \toprule
         Train Subset   & Method & BigGAN & Midjourney & Wukong & SD V1.4 & SD V1.5 & ADM & GLIDE & VQDM & Avg. \\
        \midrule
        \multirow{3}{*}{BigGAN} 
            & LOTA & 100 & 91.3 & 99.8 & 99.9 & 99.9 & 99.4 & 98.2 & 99.4 &  98.6 \\
            & LOTA-\textit{scl.} & 100 & 91.4 & 99.8 & 99.9 & 99.9 & 99.5 & 98.3 & 99.5 & 98.6 \\
            & LOTA-\textit{ngc} & 100 & 82.0 & 63.5 & 62.2 & 62.7 & 84.8 & 83.8 & 88.4 & 76.3 \\
        \midrule
        \multirow{3}{*}{Midjourney} 
            & LOTA & 98.9 & 98.8 & 99.0 & 99.0 & 98.9 & 99.0 & 99.0 & 99.0 & 99.0 \\
            & LOTA-\textit{scl.} & 98.9 & 98.8 & 98.1 & 98.9 & 98.9 & 99.0 & 99.1 & 98.0 & 98.9 \\
            & LOTA-\textit{ngc} & 97.4 & 99.7 & 99.7 & 99.7 & 99.7 & 99.8 & 99.3 & 99.3 & 98.6 \\
        \midrule
        \multirow{3}{*}{Wukong} 
            & LOTA & 100 & 93.2 & 100 & 99.9 & 100 & 99.8 & 99.9 & 99.7 & 99.1 \\
            & LOTA-\textit{scl.} & 100 & 91.6 & 99.8 & 99.9 & 99.9 & 99.5 & 98.3 & 99.5 & 98.6 \\
            & LOTA-\textit{ngc} & 85.7 & 91.3 & 99.9 & 100 & 100 & 100 & 100 & 100 & 97.9 \\
        \midrule
        \multirow{3}{*}{SD V1.4} 
            & LOTA & 100 & 91.3 & 99.8 & 99.9 & 99.9 & 99.4 & 98.2 & 99.4 & 98.5 \\
            & LOTA-\textit{scl.} & 100 & 91.6 & 99.8 & 99.9 & 99.9 & 99.5 & 98.3 & 99.5 & 98.6 \\
            & LOTA-\textit{ngc} & 70.4 & 93.6 & 100 & 100 & 100 & 100 & 100 & 100 & 95.6 \\
        \midrule
        \multirow{3}{*}{SD V1.5} 
            & LOTA & 100 & 93.1 & 100 & 100 & 100 & 99.7 & 100 & 99.7 & 99.1 \\
            & LOTA-\textit{scl.} & 100 & 91.6 & 99.8 & 99.9 & 99.9 & 99.5 & 98.3 & 99.5 & 98.6 \\
            & LOTA-\textit{ngc} & 94.9 & 92.3 & 100 & 100 & 100 & 100 & 100 & 100 & 96.8 \\
        \midrule
        \multirow{3}{*}{ADM} 
            & LOTA & 100 & 91.2 & 99.8 & 99.9 & 99.9 & 99.4 & 98.2 & 99.4 & 98.5 \\
            & LOTA-\textit{scl.} & 100 & 91.6 & 99.8 & 99.9 & 99.9 & 99.5 & 98.4 & 99.5 & 98.6 \\
            & LOTA-\textit{ngc} & 71.1 & 91.0 & 99.9 & 99.9 & 99.9 & 100 & 100 & 100 & 95.2 \\
        \midrule
        \multirow{3}{*}{GLIDE} 
            & LOTA & 100 & 94.0 & 100 & 100 & 100 & 99.8 & 100 & 99.8 & 99.2 \\
            & LOTA-\textit{scl.} & 99.2 & 96.4 & 99.3 & 99.2 & 99.1 & 99.2 & 99.2 & 99.2 & 98.9 \\
            & LOTA-\textit{ngc} & 90.5 & 85.6 & 92.0 & 91.2 & 91.1 & 99.3 & 99.2 & 99.4 & 92.1 \\
        \midrule
        \multirow{3}{*}{VQDM} 
            & LOTA & 99.9 & 92.7 & 100 & 99.9 & 100 & 99.6 & 99.8 & 99.6 & 99.0 \\
            & LOTA-\textit{scl.} & 100 & 91.6 & 99.8 & 99.9 & 99.9 & 99.5 & 98.3 & 99.5 & 98.6 \\
            & LOTA-\textit{ngc} & 94.5 & 87.3 & 91.9 & 92.3 & 92.8 & 100 & 100 & 100 & 92.7 \\
        \bottomrule
    \end{tabular}
    }
\end{table*}

\begin{figure*}[h]
  \centering
  \small 
\includegraphics[width=1\linewidth]{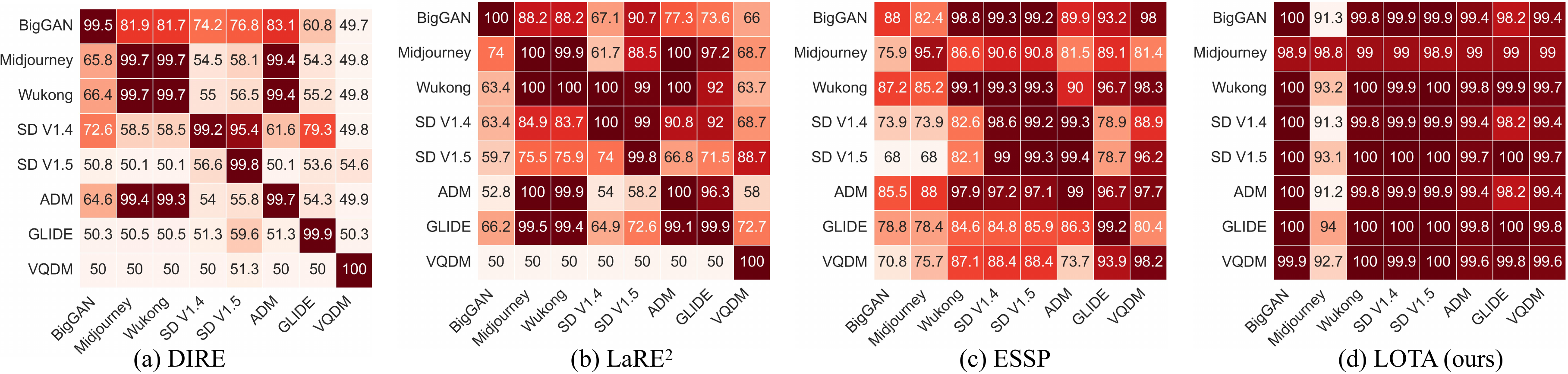}
  \caption{\textbf{Comparison of cross-generator generalization with existing methods.} DIRE \cite{wang2023dire}, LaRE$^2$ \cite{luo2024lare} and ESSP \cite{chen2024single} are selected as comparison methods. These models and ours are trained on eight training subsets and tested on eight testing subsets in the GenImage. The noise-based classifier is applied on both training and testing. The vertical axis represents the training subsets, and the horizontal axis represents the test subsets. Darker colors indicate higher accuracy.}
  \label{fig:heatmap}
\end{figure*}

\noindent\textbf{Analysis of Experimental Results:}
We train our model on eight training subsets of GenImage and evaluate each trained model on all eight testing subsets in Table~\ref{tab:experiments}. The default LOTA uses thresholding when generating low-bit images and employs noise-based classification. The two other variants, LOTA-\textit{scl.} and LOTA-\textit{ngc}, use scaling and noise-guided classifier, respectively. We compare the noise image generation approaches of scaling and thresholding, and find that thresholding achieves slightly higher average accuracy compared to scaling. From Figure~\ref{fig:heatmap}, we find that results of the three subsets (\textit{e.g}., Wukong, Stable Diffusion V1.4, and Stable Diffusion V1.5) are highly correlated, which may be attributed to the high correlation of the generators.




\begin{table*}[h]
    \centering
    \caption{\textbf{Comparison of averaged accuracy against existing methods on the GenImage dataset.} Models are trained and tested on eight subsets of GenImage, and the average accuracy is reported. LOTA-\textit{scl.} and LOTA-\textit{ngc} denote the variants that use scaling and noise-guided classifier, respectively. `*' means the results are reproduced by ourselves.}
    \vspace{-0.5em}
    \label{tab:comparison}
    
    \adjustbox{max width=0.92\textwidth}{
    \begin{tabular}{lC{1.8cm}C{1.4cm}C{1.4cm}C{1.4cm}C{1.4cm}C{1.4cm}C{1.4cm}C{1.4cm}C{1.4cm}C{1.4cm}}
        \toprule
        Method  & Error-Based & BigGAN & Midjourney & Wukong & SD V1.4 & SD V1.5 & ADM & GLIDE & VQDM & Avg. \\
        \midrule
        CNNSpot \cite{wang2020cnn} & \ding{55} & 56.6 & 58.2 & 67.7 & 70.3 & 70.2 & 57.0 & 57.1 & 56.7 & 61.7 \\
        F3Net \cite{qian2020thinkingF3Net} & \ding{55} & 56.5 & 55.1 & 72.3 & 73.1 & 73.1 & 66.5 & 57.8 & 62.1 & 64.6 \\
        GramNet \cite{liu2020global} & \ding{55} & 61.2 & 58.1 & 71.3 & 72.8 & 72.7 & 58.7 & 65.3 & 57.8 & 64.7 \\
        Spec \cite{zhang2019detectingSPEC} & \ding{55} & 64.3 & 56.7 & 70.3 & 72.4 & 72.3 & 57.9 & 65.4 & 61.7 & 65.1 \\
        ResNet-50 \cite{he2016deepresnet} & \ding{55} & 66.6 & 59.0 & 71.4 & 72.3 & 72.4 & 59.7 & 73.1 & 60.9 & 66.9 \\
        DeiT-S \cite{touvron2021training} & \ding{55} & 66.3 & 60.7 & 73.1 & 74.2 & 74.2 & 59.5 & 71.1 & 61.7 & 67.6 \\
        Swin-T \cite{liu2021swin} & \ding{55} & 69.5 & 61.7 & 75.1 & 76.0 & 76.1 & 61.3 & 76.9 & 65.8 & 70.3 \\
        \midrule
        DIRE* \cite{wang2023dire} & \ding{51} & 56.7 & 59.7 & 74.6 & 74.7 & 74.7 & 68.8 & 69.3 & 68.8 & 73.4 \\
        LaRE$^2$* \cite{luo2024lare} & \ding{51} & 74.0 & 66.4 & 85.5 & 87.5 & 87.3 & 66.6 & 81.3 & 84.4 & 79.4 \\
        ESSP* \cite{chen2024single} & \ding{51} & 78.3 & 80.8 & 93.5 & 94.2 & 84.4 & 82.1 & 92.1 & 91.0 & 87.0 \\
        \midrule
        LOTA-\textit{scl.} & \ding{51} & 99.8 & 93.1 & 99.5 & 99.7 & 99.7 & 99.4 & 98.5 & 99.3 & 98.7 \\
        LOTA-\textit{ngc} & \ding{51} & 88.1 & 90.4 & 93.4 & 93.2 & 93.3 & 98.0 & 97.8 & 98.4 & 93.2 \\
        LOTA & \ding{51} & \textbf{99.9} & \textbf{93.2} & \textbf{99.8} & \textbf{99.8} & \textbf{99.8} & \textbf{99.5} & \textbf{99.2} & \textbf{99.5} & \textbf{98.9} \\
        \bottomrule
    \end{tabular}}
\end{table*}

\noindent\textbf{Comparison with State-of-the-Arts:} 
In Table \ref{tab:comparison}, the proposed LOTA achieves an average accuracy of 98.9\%, while its two variants, LOTA-\textit{scl.} and LOTA-\textit{ngc}, achieve 98.7\% and 93.2\%, respectively. All three significantly outperform current mainstream methods. Specifically, for error extraction based methods, LOTA shows improvements of approximately 11.9\% over ESSP \cite{chen2024single}, 19.6\% over LaRE$^2$ \cite{luo2024lare}, and 25.5\% over DIRE \cite{wang2023dire}.

\noindent\textbf{Comparison of Cross-Generator Generalization:} Results of cross-generator generalization are compared in Figure~\ref{fig:heatmap}. Existing methods based on error extraction predominantly exhibit darker colors along the main diagonal, indicating their optimal performance only when training and testing generators are identical. While LaRE$^2$ \cite{luo2024lare} and ESSP \cite{chen2024single} show improved cross-generator generalization with darker off-diagonal entries, their capability remains confined to homologous generators (e.g., Diffusion model generators including Stable Diffusion V1.4 \cite{rombach2022high}, Stable Diffusion V1.5 \cite{rombach2022high}, ADM \cite{dhariwal2021diffusion}, and GLIDE \cite{nichol2021glide}). In contrast, LOTA demonstrates uniformly distributed color patterns across all rows and columns, reflecting its superior cross-generator generalization. Notably, even the lightest-colored column representing Midjourney achieves higher detection accuracy than results from other mainstream methods.

\begin{table*}[ht]
    \centering
    \begin{minipage}[t]{\columnwidth} 
        \centering
        \caption{\textbf{Ablation.} We conduct ablation studies by removing the modules of BGNIG, MGPS and NBC/NGC, respectively.}
         \vspace{-0.2cm}
        \adjustbox{max width=\textwidth}{
        \begin{tabular}{lC{0.45cm}C{0.45cm}C{0.45cm}C{0.45cm}C{0.45cm}C{0.45cm}C{0.45cm}C{0.45cm}C{0.45cm}}
            \toprule
        Ablations & Big & Mid & Wuk & SD4 & SD5 & ADM & GLI & VQD & Avg.\\
        \midrule
        LOTA-\textit{ngc}   & \textbf{94.9} & \textbf{92.3} & \textbf{100} & \textbf{100} & \textbf{100} & \textbf{100} & \textbf{100} & \textbf{100} & \textbf{96.8} \\
        w/o BGNIG & 77.6 & 92.8 & 92.5 & 80.5 & 99.8 & 99.8 & 88.7 & 84.3 & 87.5 \\
        w/o MGPS & 82.2 & 81.1 & 94.6 & 96.8 & 96.6 & 77.3 & 96.7 & 99.9 & 90.9 \\
        w/o NBC/NGC & 52.9 & 52.4 & 58.1 & 55.1 & 55.2 & 50.5 & 51.7 & 50.8 & 52.6 \\
        \bottomrule
        \end{tabular}
        }
       \label{tab:ablation}
    \end{minipage}
    \hfill
    \begin{minipage}[t]{\columnwidth} 
                \centering
        \caption{\textbf{Impact of Patch Size.} We evaluate different size of patch, e.g., 16$\times$16, 32$\times$32, 48$\times$48, and 64$\times$64, for the MGPS module.}
         \vspace{-0.2cm}
        \adjustbox{max width=\textwidth}{
        \begin{tabular}{lC{0.45cm}C{0.45cm}C{0.45cm}C{0.45cm}C{0.45cm}C{0.45cm}C{0.45cm}C{0.45cm}C{0.45cm}}
            \toprule
            Patch Size & Big & Mid & Wuk & SD4 & SD5 & ADM & GLI & VQD & Avg.\\
            \midrule
            16 $\times$ 16 & 50.2 & 95.7 & 99.9 & 99.9 & 99.9 & 99.9 & 99.9 & 99.9 & 92.7 \\
            32 $\times$ 32 & \textbf{94.9} & \textbf{92.3} & \textbf{100} & \textbf{100} & \textbf{100} & \textbf{100} & \textbf{100} & \textbf{100} & \textbf{96.8} \\
            48 $\times$ 48 & 83.8 & 78.0 & 100 & 100 & 100 & 100 & 100 & 100 & 91.4 \\
            64 $\times$ 64 & 53.6 & 84.0 & 100 & 100 & 99.9 & 100 & 100 & 100 & 90.8 \\
        \bottomrule
        \end{tabular}
        }
        \label{tab:patch_size}
    \end{minipage}
    \begin{minipage}[t]{\columnwidth} 
                \centering
        \vspace{0.2cm}
        \caption{\textbf{Impact of Patch Selection Strategy.} We select patches randomly, by highest score, or by lowest score.}
         \vspace{-0.2cm}
        \adjustbox{max width=\textwidth}{
        \begin{tabular}{lC{0.45cm}C{0.45cm}C{0.45cm}C{0.45cm}C{0.45cm}C{0.45cm}C{0.45cm}C{0.45cm}C{0.45cm}}
            \toprule
            Selection & Big & Mid & Wuk & SD4 & SD5 & ADM & GLI & VQD & Avg.\\
            \midrule
        Random  & 69.9 & 98.1 & 99.8 & 99.8 & 99.8 & 99.4 & 98.6 & 99.4 & 93.1 \\
        Min     & 79.0 & 91.9 & 99.0 & 99.2 & 99.2 & 96.1 & 95.3 & 96.3 & 89.6 \\
        Max     & \textbf{94.9} & \textbf{92.3} & \textbf{100} & \textbf{100} & \textbf{100} & \textbf{100} & \textbf{100} & \textbf{100} & \textbf{96.8} \\
        \bottomrule
        \end{tabular}
        }
        \label{tab:patch_selection}
    \end{minipage}
    \hfill
     \begin{minipage}[t]{\columnwidth} 
                \centering
        \vspace{0.2cm}
        \caption{\textbf{Impact of Classifier.} We choose a simple FC layer, noise-based and noise-guided classifier as the final classifier.}
        \vspace{-0.2cm}
        \adjustbox{max width=\textwidth}{
        \begin{tabular}{lC{0.45cm}C{0.45cm}C{0.45cm}C{0.45cm}C{0.45cm}C{0.45cm}C{0.45cm}C{0.45cm}C{0.45cm}}
            \toprule
            Classifier & Big & Mid & Wuk & SD4 & SD5 & ADM & GLI & VQD & Avg.\\
            \midrule
           FC    & 52.9 & 52.4 & 58.1 & 55.1 & 55.2 & 50.5 & 51.7 & 50.8 & 52.6 \\
        NBC & \textbf{100} & \textbf{93.1} & \textbf{100} & \textbf{100} & \textbf{100} & \textbf{99.7} & \textbf{100} & \textbf{99.7} & \textbf{99.1} \\
        NGC & 94.9 & 92.3 & 100 & 100 & 100 & 100 & 100 & 100 & 96.8 \\
        \bottomrule
        \end{tabular}
        }
        \label{tab:classifier}
    \end{minipage}
\end{table*}

\subsection{Ablation Studies and Analysis}
We conduct ablation studies to validate the effectiveness of each module. We train models on the Stable Diffusion V1.5 \cite{rombach2022high} subset and test on all 8 subsets. By default, we use the LOTA variant with a noise-guided classifier as our baseline, denoted as LOTA-\textit{ngc}. For simplicity, the subsets of BigGAN, Midjourney, Wukong, SD V1.4, SD V1.5, ADM, GLIDE, and VQDM are abbreviated as Big, Mid, Wuk, SD4, SD5, ADM, GLI, and VQD, respectively. 

\noindent \textbf{Ablation Studies:}
To validate the effectiveness of each module, we respectively remove each module proposed in our method: 1) w/o BGNIG: we employ raw images instead of low-bit images for detection. 2) w/o MGPS: we do not crop images into patches, and use images of original size. 3) w/o NBC/NGC: we replace the noise-based or noise-guided classifier with a fully-connected layer. As shown in Table~\ref{tab:ablation}, it exhibits a clear accuracy decline (especially testing on BigGAN and Midjourney) when using raw images only, which demonstrates that low-bit planes extraction effectively exploits the intrinsic patterns of images and supplements the capability of cross-generator generalization. We also find there is a significant accuracy drop when NBC and NGC classifier are discarded, indicating the feature refinement capability of our proposed classifier.

\noindent \textbf{Impact of Patch Size:}
The selection of patch size is critical for further detection, as a too large patch would introduce texture or useless information, and a too small patch would weaken the noise pattern for detection. So we vary the selection of patch size, including 16$\times$16, 32$\times$32, 48$\times$48, and 64$\times$64. As shown in Table~\ref{tab:patch_size}, as the patch size increases, the average accuracy first increases and then decreases. When using the patch of size 32$\times$32, the average accuracy is much higher than other cases. Apart from this, we also find that fluctuations in average accuracy are due to variations in BigGAN subset (a heterologous generator), so the patch size also influences the cross-generator generalization. So we choose the patch of size 32$\times$32 to maximize the cross-generator generalization.

\noindent \textbf{Impact of Patch Selection Strategy:}
In the MGPS module,  we choose the patch with the highest gradient-based score for subsequent processes based on the assumption that higher score indicates greater high-frequency variations for real images or more obvious artifacts for generated images. To validate this, we modify the selection method by using the maximum score, minimum score, and random selection. As shown in Table~\ref{tab:patch_selection}, the selection based on the minimal score gets the lowest accuracy, while the selection based on the maximum score gets the highest accuracy. 
Consequently, we choose the patch with the highest score to exploit the intrinsic patterns of noisy images.

\noindent \textbf{Impact of Classifier:}
We compare our Noise-Based Classifier (NBC) and Noise-Guided Classifier (NGC) with a simple Fully Connected (FC) layer, where the error map is directly sent to a FC layer. As shown in Table~\ref{tab:classifier}, FC itself cannot effectively extract useful information from low-bit images, achieving an average ACC of only 52.6\%. In contrast, the NBC and NGC classifiers fully leverage the noise patterns in low-bit images to distinguish AI-generated images from real ones, attaining average ACCs of 99.1\% and 96.8\%, respectively. 

\noindent \textbf{Impact of the Number of Bit-Planes:}
We choose the three least  bit-planes to generate the noise image based on the assumption that lower-order bit-planes preserve more noise patterns, which are the critical indicator for distinguishing AI-generated images from real ones. To validate this point, we choose different numbers of bits to generate an image: from 0-bit to 5-bit. As shown in Table~\ref{tab:bit-planes}, as the number of combined planes increases, the average accuracy firstly increases to 96.8\% when the three lowest bit-planes are composed, and then decreases. Fluctuation is especially obvious when testing on the subsets of BigGAN and Midjourney. This suggests that lower bit-planes contain less excessive high-frequency variation, making it difficult to detect key features, while higher bit-planes introduce some visual features that obscure certain noise patterns, thereby impairing the cross-generator generalization capability.

\begin{figure}[tb]
    \centering
    \begin{subfigure}{0.23\textwidth}
        \centering
        \includegraphics[width=\linewidth]{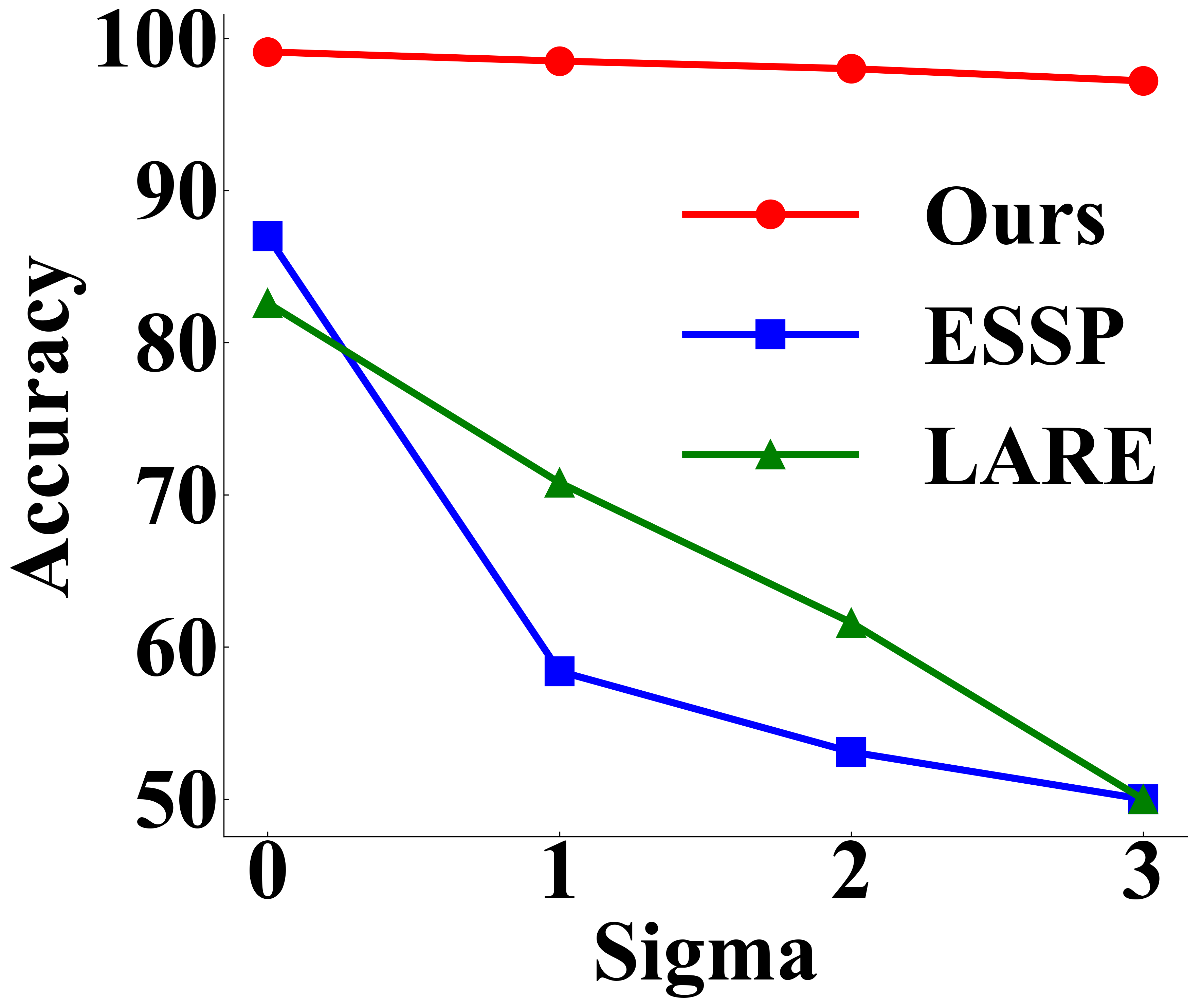}
        \caption{Gaussian Blur}
        \label{fig:gaussian}
    \end{subfigure}
    \begin{subfigure}{0.23\textwidth}
        \centering
        \includegraphics[width=\linewidth]{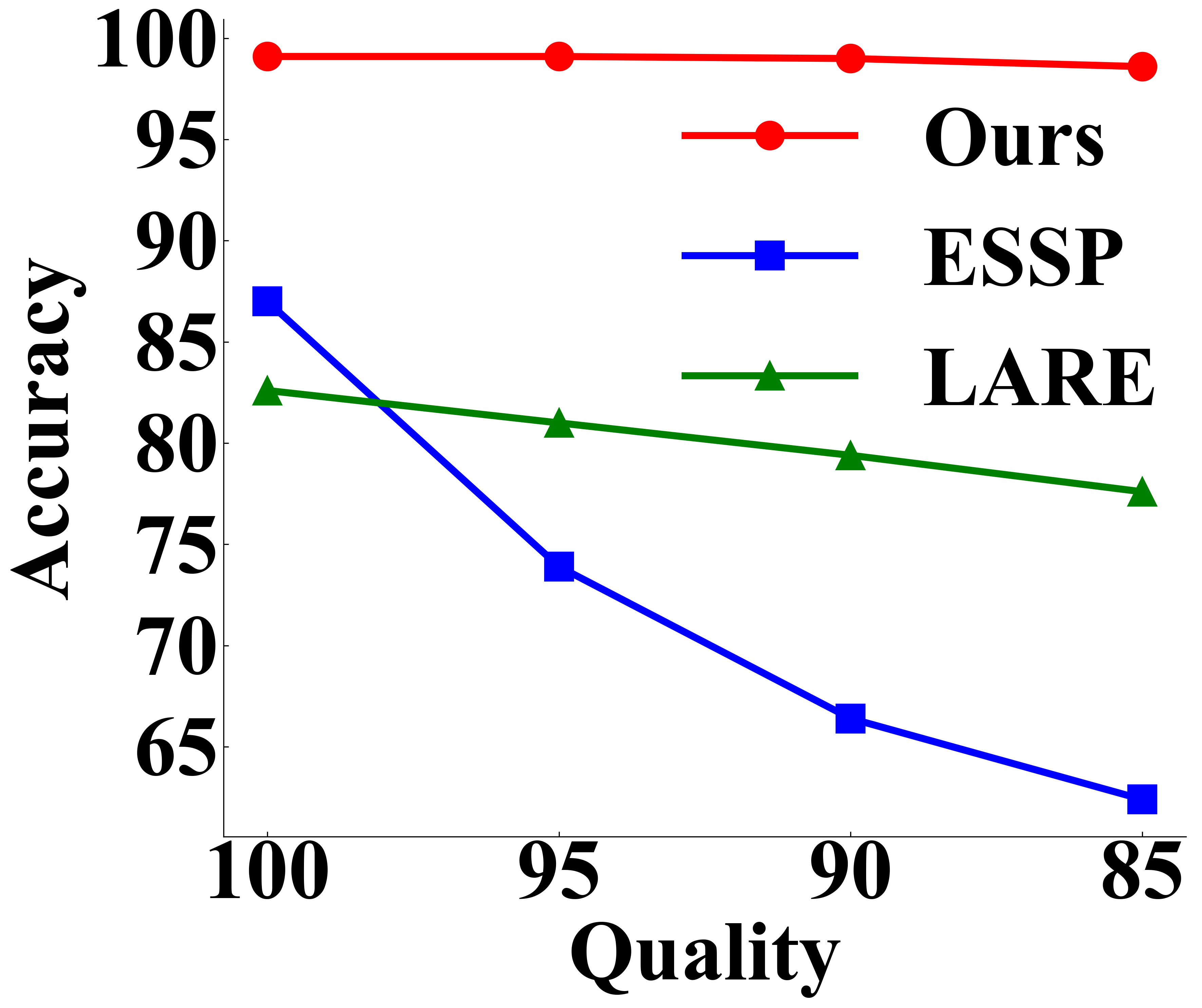}
        \caption{JPEG Compression}
        \label{fig:jpeg}
    \end{subfigure}
    \caption{\textbf{Robustness to Image Degradation.} Gaussian blur with $\sigma$ = 0,1,2,3 and JPEG compression (\textit{quality} = 100\%,95\%,90\%,85\%) are applied to LaRE$^2$ \cite{luo2024lare}, ESSP \cite{chen2024single} and our LOTA, the results demonstrate the robustness of ours to unseen perturbations.}
    \label{fig:robustness}
\end{figure}

\begin{table}[tb]
    \small
    \centering
    \caption{\textbf{Impact of Bit-Planes.} We generate the noise image by combining different numbers of bit-planes, ranging from 0 to 5.}
    \label{tab:bit-planes}
    \resizebox{\linewidth}{!}{
    \begin{tabular}{lC{0.45cm}C{0.45cm}C{0.45cm}C{0.45cm}C{0.45cm}C{0.45cm}C{0.45cm}C{0.45cm}C{0.45cm}}
       \toprule
            Bit-Planes   & Big & Mid & Wuk & SD4 & SD5 & ADM & GLI & VQD & Avg.\\
            \midrule
            0     & 78.8 & 88.3 & 100 & 100 & 100 & 99.8 & 99.4 & 100 & 91.5 \\
            0$\sim$1 & 93.6 & 77.0 & 100 & 100 & 100 & 99.9 & 99.9 & 100 & 95.1 \\
            0$\sim$2 & \textbf{94.9} & \textbf{92.3} & \textbf{100} & \textbf{100} & \textbf{100} & \textbf{100} & \textbf{100} & \textbf{100} & \textbf{96.8} \\
            0$\sim$3 & 85.5 & 97.0 & 99.9 & 100 & 99.9 & 100 & 100 & 100 & 95.7 \\
            0$\sim$4 & 89.2 & 96.3 & 99.9 & 100 & 99.9 & 99.9 & 100 & 100 & 93.4 \\
            0$\sim$5 & 91.4 & 96.1 & 98.8 & 99.3 & 99.1 & 85.6 & 82.6 & 79.0 & 86.4 \\
    \bottomrule
    \end{tabular}
    }
\end{table}



\noindent \textbf{Analysis of Computational Efficiency}: For time consumption of error extraction and deepfake image classification, DIRE \cite{wang2023dire} uses 20 steps to build an error map, totally consuming 2 seconds per image, and LaRE$^2$ \cite{luo2024lare} uses 1 step, totally consuming 0.26 seconds per image. Even the existing fastest ESSP \cite{chen2024single} consumes 31.99 milliseconds to process an image. As shown in Table~\ref{tab:comparison_time}, leveraging bit-plane based operations, LOTA uses only one step to generate an error image, which operates at the millisecond level (1.52 milliseconds for error extraction and 4.00 milliseconds in total), demonstrating nearly a hundred times faster than existing methods in error extraction. Regarding the number of parameters, several mainstream methods rely on large pre-trained models (e.g., Diffusion), resulting in models with abundant parameters. Our model has 23.6M parameters for deepfake image detection, a breakthrough that significantly enhances its practical deployment potential.

\begin{table}[t]
    \small
    \centering
    \caption{\textbf{Comparison in Computation Efficiency.} We compare our methods with different classifiers with other mainstream methods based on error extraction, considering two dimensions:  1) Time consumption for error extraction and classification—our method operates at the millisecond level, nearly a hundred times faster than other methods. 2) Model parameters—our method is efficient and lightweight, requiring significantly fewer parameters than methods that rely on large pre-trained models, which introduce substantial computational overhead.}
    \label{tab:comparison_time}
     \resizebox{0.9\linewidth}{!}{
    \begin{tabular}{lcccc}
        \toprule
        \multirow{3}{*}{Method} & \multicolumn{2}{c}{Time} & \multicolumn{2}{c}{Params} \\
        \cmidrule(lr){2-3} \cmidrule(lr){4-5}
        & {\makecell{Error\\Extraction}} & Total & {\makecell{Error\\Extraction}} & Total\\
        \midrule
        DIRE \cite{wang2023dire}        & 1.99 s   & 2 s   & 644.8M  & 688.3M \\
        LaRE$^2$ \cite{luo2024lare}     & 250 ms   & 260 ms   & 1066.2M & 1165.8M \\
        ESSP \cite{chen2024single}      & 25.10 ms & 31.99 ms & 7.1M    & 30.7M \\
        \midrule
        LOTA-NBC                         & 1.52 ms  & 4.00 ms  & 0      & 23.6M \\
        LOTA-NGC                         & 1.52 ms  & 4.71 ms  & 0      & 28.4M \\
        \bottomrule
    \end{tabular}}
\end{table}

\noindent \textbf{Robustness to Image Degradation}: To further demonstrate the robustness to unseen perturbations and degradations, we train our model and comparison models on Stable Diffusion V1.5 \cite{rombach2022high}, and test on 8 subsets which are processed by Gaussian blur and JPEG compression to different extents. Following \cite{wang2020cnn} and \cite{wang2023dire}, we choose Gaussian blur ($\sigma$ = 0,1,2,3) and JPEG compression (\textit{quality} = 100\%,95\%,90\%,85\%) when testing, the results are shown in Figure~\ref{fig:robustness}. We observe that when the image degrades, existing methods, like LaRE$^2$ \cite{luo2024lare} and ESSP \cite{chen2024single}, encounter a huge decline. Especially when the $\sigma$ of the Gaussian blur is set to 2 and 3, the two models nearly degenerate into random guess classifiers. In contrast, our method is very stable against these interference, demonstrating the strong robustness to unseen perturbations and degradations.


%% file: sec/5_conclusion.tex
\section{Conclusion}
\label{sec:conclusion}
In this paper, we propose an effective and efficient AI-generated image detection method called LOTA. LOTA innovatively leverages bit-planes for noisy image generation and fake detection. It comprises three decoupled and concise modules: Bit-plane Guided Noisy Image Generation (BGNIG), Maximum Gradient Patch Selection (MGPS) and classification head. Extensive experiments on the GenImage benchmark demonstrate the outstanding performance and strong cross-generator generalization capability of our method. Our approach is highly robust to unseen perturbations and degradations. It contains only 23.6 million parameters and operates at the millisecond level, making it nearly a hundred times faster than existing methods.

%% file: main.bbl
\begin{thebibliography}{42}
\providecommand{\natexlab}[1]{#1}
\providecommand{\url}[1]{\texttt{#1}}
\expandafter\ifx\csname urlstyle\endcsname\relax
  \providecommand{\doi}[1]{doi: #1}\else
  \providecommand{\doi}{doi: \begingroup \urlstyle{rm}\Url}\fi

\bibitem[Mid(2022)]{Midjourney}
Midjourney.
\newblock \url{https://www.midjourney.com/home/}, 2022.

\bibitem[wuk(2022)]{wukong}
Wukong.
\newblock \url{https://xihe.mindspore.cn/modelzoo/wukong}, 2022.

\bibitem[Brock et~al.(2018)]{BigGAN}
Andrew Brock et~al.
\newblock Large scale gan training for high fidelity natural image synthesis.
\newblock In \emph{ICLR}, 2018.

\bibitem[Cao et~al.(2025)Cao, Wu, Cao, Liu, and Gui]{cao2025external}
Biwei Cao, Qihang Wu, Jiuxin Cao, Bo Liu, and Jie Gui.
\newblock External reliable information-enhanced multimodal contrastive
  learning for fake news detection.
\newblock In \emph{AAAI}, pages 31--39, 2025.

\bibitem[Chandrasegaran et~al.(2021)Chandrasegaran, Tran, and
  Cheung]{chandrasegaran2021closer}
Keshigeyan Chandrasegaran, Ngoc-Trung Tran, and Ngai-Man Cheung.
\newblock A closer look at fourier spectrum discrepancies for cnn-generated
  images detection.
\newblock In \emph{CVPR}, pages 7200--7209, 2021.

\bibitem[Chen et~al.(2024{\natexlab{a}})Chen, Zeng, Yang, and
  Yang]{chen2024drct}
Baoying Chen, Jishen Zeng, Jianquan Yang, and Rui Yang.
\newblock Drct: Diffusion reconstruction contrastive training towards universal
  detection of diffusion generated images.
\newblock In \emph{ICML}, 2024{\natexlab{a}}.

\bibitem[Chen et~al.(2024{\natexlab{b}})Chen, Yao, and Niu]{chen2024single}
Jiaxuan Chen, Jieteng Yao, and Li Niu.
\newblock A single simple patch is all you need for ai-generated image
  detection.
\newblock \emph{arXiv preprint arXiv:2402.01123}, 2024{\natexlab{b}}.

\bibitem[Corvi et~al.(2023{\natexlab{a}})Corvi, Cozzolino, Poggi, Nagano, and
  Verdoliva]{corvi2023intriguing}
Riccardo Corvi, Davide Cozzolino, Giovanni Poggi, Koki Nagano, and Luisa
  Verdoliva.
\newblock Intriguing properties of synthetic images: from generative
  adversarial networks to diffusion models.
\newblock In \emph{CVPR}, pages 973--982, 2023{\natexlab{a}}.

\bibitem[Corvi et~al.(2023{\natexlab{b}})Corvi, Cozzolino, Zingarini, Poggi,
  Nagano, and Verdoliva]{corvi2023detection}
Riccardo Corvi, Davide Cozzolino, Giada Zingarini, Giovanni Poggi, Koki Nagano,
  and Luisa Verdoliva.
\newblock On the detection of synthetic images generated by diffusion models.
\newblock In \emph{ICASSP}, pages 1--5. IEEE, 2023{\natexlab{b}}.

\bibitem[Cozzolino et~al.(2024)Cozzolino, Poggi, Nie{\ss}ner, and
  Verdoliva]{cozzolino2024zero}
Davide Cozzolino, Giovanni Poggi, Matthias Nie{\ss}ner, and Luisa Verdoliva.
\newblock Zero-shot detection of ai-generated images.
\newblock In \emph{ECCV}, pages 54--72. Springer, 2024.

\bibitem[Dhariwal and Nichol(2021)]{dhariwal2021diffusion}
Prafulla Dhariwal and Alexander Nichol.
\newblock Diffusion models beat gans on image synthesis.
\newblock \emph{NeurIPS}, 2021.

\bibitem[Doloriel and Cheung(2024)]{doloriel2024frequency}
Chandler~Timm Doloriel and Ngai-Man Cheung.
\newblock Frequency masking for universal deepfake detection.
\newblock In \emph{ICASSP}, pages 13466--13470. IEEE, 2024.

\bibitem[Dzanic et~al.(2020)Dzanic, Shah, and Witherden]{dzanic2020fourier}
Tarik Dzanic, Karan Shah, and Freddie Witherden.
\newblock Fourier spectrum discrepancies in deep network generated images.
\newblock In \emph{NeurIPS}, pages 3022--3032, 2020.

\bibitem[Elharrouss et~al.(2020)Elharrouss, Almaadeed, and
  Al-Maadeed]{elharrouss2020image}
Omar Elharrouss, Noor Almaadeed, and Somaya Al-Maadeed.
\newblock An image steganography approach based on k-least significant bits
  (k-lsb).
\newblock In \emph{IEEE international conference on informatics, IoT, and
  enabling technologies (ICIoT)}, pages 131--135. IEEE, 2020.

\bibitem[Goodfellow et~al.(2014)Goodfellow, Pouget-Abadie, Mirza, Xu,
  Warde-Farley, Ozair, Courville, and Bengio]{goodfellow2014generative}
Ian Goodfellow, Jean Pouget-Abadie, Mehdi Mirza, Bing Xu, David Warde-Farley,
  Sherjil Ozair, Aaron Courville, and Yoshua Bengio.
\newblock Generative adversarial nets.
\newblock In \emph{NIPS}, 2014.

\bibitem[Gu et~al.(2022)Gu, Chen, Bao, Wen, Zhang, Chen, Yuan, and
  Guo]{gu2022vectorvqdm}
Shuyang Gu, Dong Chen, Jianmin Bao, Fang Wen, Bo Zhang, Dongdong Chen, Lu Yuan,
  and Baining Guo.
\newblock Vector quantized diffusion model for text-to-image synthesis.
\newblock In \emph{CVPR}, pages 10696--10706, 2022.

\bibitem[He et~al.(2016)He, Zhang, Ren, and Sun]{he2016deepresnet}
Kaiming He, Xiangyu Zhang, Shaoqing Ren, and Jian Sun.
\newblock Deep residual learning for image recognition.
\newblock In \emph{CVPR}, pages 770--778, 2016.

\bibitem[Johnson and Jajodia(1998)]{johnson1998exploring}
Neil~F Johnson and Sushil Jajodia.
\newblock Exploring steganography: Seeing the unseen.
\newblock \emph{Computer}, 31\penalty0 (2):\penalty0 26--34, 1998.

\bibitem[Ju et~al.(2023)Ju, Jia, Cai, Guan, and Lyu]{ju2023glff}
Yan Ju, Shan Jia, Jialing Cai, Haiying Guan, and Siwei Lyu.
\newblock Glff: Global and local feature fusion for ai-synthesized image
  detection.
\newblock \emph{IEEE Transactions on Multimedia}, 26:\penalty0 4073--4085,
  2023.

\bibitem[Juefei-Xu et~al.(2022)Juefei-Xu, Wang, Huang, Guo, Ma, and
  Liu]{juefei2022countering}
Felix Juefei-Xu, Run Wang, Yihao Huang, Qing Guo, Lei Ma, and Yang Liu.
\newblock Countering malicious deepfakes: Survey, battleground, and horizon.
\newblock \emph{IJCV}, 2022.

\bibitem[Kawaguchi and Eason(1999)]{kawaguchi1999principles}
Eiji Kawaguchi and Richard~O Eason.
\newblock Principles and applications of bpcs steganography.
\newblock In \emph{Multimedia systems and applications}, pages 464--473. SPIE,
  1999.

\bibitem[Lanzino et~al.(2024)Lanzino, Fontana, Diko, Marini, and
  Cinque]{lanzino2024faster}
Romeo Lanzino, Federico Fontana, Anxhelo Diko, Marco~Raoul Marini, and Luigi
  Cinque.
\newblock Faster than lies: Real-time deepfake detection using binary neural
  networks.
\newblock In \emph{CVPR}, pages 3771--3780, 2024.

\bibitem[Leporoni et~al.(2024)Leporoni, Maiano, Papa, and
  Amerini]{leporoni2024guided}
Giorgio Leporoni, Luca Maiano, Lorenzo Papa, and Irene Amerini.
\newblock A guided-based approach for deepfake detection: Rgb-depth integration
  via features fusion.
\newblock \emph{Pattern Recognition Letters}, 181:\penalty0 99--105, 2024.

\bibitem[Liu et~al.(2022)Liu, Yang, Bi, Xiao, Li, and Gao]{LiuLNP}
Bo Liu, Fan Yang, Xiuli Bi, Bin Xiao, Weisheng Li, and Xinbo Gao.
\newblock Detecting generated images by real images.
\newblock In \emph{ECCV}, 2022.

\bibitem[Liu et~al.(2020)Liu, Qi, and Torr]{liu2020global}
Zhengzhe Liu, Xiaojuan Qi, and Philip~HS Torr.
\newblock Global texture enhancement for fake face detection in the wild.
\newblock In \emph{CVPR}, pages 8060--8069, 2020.

\bibitem[Liu et~al.(2021)Liu, Lin, Cao, Hu, Wei, Zhang, Lin, and
  Guo]{liu2021swin}
Ze Liu, Yutong Lin, Yue Cao, Han Hu, Yixuan Wei, Zheng Zhang, Stephen Lin, and
  Baining Guo.
\newblock Swin transformer: Hierarchical vision transformer using shifted
  windows.
\newblock In \emph{ICCV}, 2021.

\bibitem[Luo et~al.(2024)Luo, Du, Yan, and Ding]{luo2024lare}
Yunpeng Luo, Junlong Du, Ke Yan, and Shouhong Ding.
\newblock Lare\^{} 2: Latent reconstruction error based method for
  diffusion-generated image detection.
\newblock In \emph{CVPR}, pages 17006--17015, 2024.

\bibitem[Ma et~al.(2023)Ma, Duan, Kong, Shi, and Xu]{ma2023exposing}
Ruipeng Ma, Jinhao Duan, Fei Kong, Xiaoshuang Shi, and Kaidi Xu.
\newblock Exposing the fake: Effective diffusion-generated images detection.
\newblock \emph{arXiv preprint arXiv:2307.06272}, 2023.

\bibitem[Nichol et~al.(2021)Nichol, Dhariwal, Ramesh, Shyam, Mishkin, McGrew,
  Sutskever, and Chen]{nichol2021glide}
Alex Nichol, Prafulla Dhariwal, Aditya Ramesh, Pranav Shyam, Pamela Mishkin,
  Bob McGrew, Ilya Sutskever, and Mark Chen.
\newblock Glide: Towards photorealistic image generation and editing with
  text-guided diffusion models.
\newblock \emph{arXiv preprint arXiv:2112.10741}, 2021.

\bibitem[Qian et~al.(2020)Qian, Yin, Sheng, Chen, and
  Shao]{qian2020thinkingF3Net}
Yuyang Qian, Guojun Yin, Lu Sheng, Zixuan Chen, and Jing Shao.
\newblock Thinking in frequency: Face forgery detection by mining
  frequency-aware clues.
\newblock In \emph{ECCV}, pages 86--103. Springer, 2020.

\bibitem[Ricker et~al.(2022)Ricker, Damm, Holz, and Fischer]{ricker2022towards}
Jonas Ricker, Simon Damm, Thorsten Holz, and Asja Fischer.
\newblock Towards the detection of diffusion model deepfakes.
\newblock \emph{arXiv preprint arXiv:2210.14571}, 2022.

\bibitem[Rombach et~al.(2022)Rombach, Blattmann, Lorenz, Esser, and
  Ommer]{rombach2022high}
Robin Rombach, Andreas Blattmann, Dominik Lorenz, Patrick Esser, and Bj{\"o}rn
  Ommer.
\newblock High-resolution image synthesis with latent diffusion models.
\newblock In \emph{CVPR}, 2022.

\bibitem[Sarkar et~al.(2024)Sarkar, Mai, Mahapatra, Lazebnik, Forsyth, and
  Bhattad]{sarkar2024shadows}
Ayush Sarkar, Hanlin Mai, Amitabh Mahapatra, Svetlana Lazebnik, David~A
  Forsyth, and Anand Bhattad.
\newblock Shadows don't lie and lines can't bend! generative models don't know
  projective geometry... for now.
\newblock In \emph{CVPR}, pages 28140--28149, 2024.

\bibitem[Tan et~al.(2023)Tan, Zhao, Wei, Gu, and Wei]{tan2023learning}
Chuangchuang Tan, Yao Zhao, Shikui Wei, Guanghua Gu, and Yunchao Wei.
\newblock Learning on gradients: Generalized artifacts representation for
  gan-generated images detection.
\newblock In \emph{CVPR}, 2023.

\bibitem[Touvron et~al.(2021)Touvron, Cord, Douze, Massa, Sablayrolles, and
  J{\'e}gou]{touvron2021training}
Hugo Touvron, Matthieu Cord, Matthijs Douze, Francisco Massa, Alexandre
  Sablayrolles, and Herv{\'e} J{\'e}gou.
\newblock Training data-efficient image transformers \& distillation through
  attention.
\newblock In \emph{ICML}, 2021.

\bibitem[Vaswani et~al.(2017)Vaswani, Shazeer, Parmar, Uszkoreit, Jones, Gomez,
  Kaiser, and Polosukhin]{vaswani2017attention}
Ashish Vaswani, Noam Shazeer, Niki Parmar, Jakob Uszkoreit, Llion Jones,
  Aidan~N Gomez, {\L}ukasz Kaiser, and Illia Polosukhin.
\newblock Attention is all you need.
\newblock In \emph{NeurIPS}, 2017.

\bibitem[Wang et~al.(2020)Wang, Wang, Zhang, Owens, and Efros]{wang2020cnn}
Sheng-Yu Wang, Oliver Wang, Richard Zhang, Andrew Owens, and Alexei~A Efros.
\newblock Cnn-generated images are surprisingly easy to spot... for now.
\newblock In \emph{CVPR}, pages 8695--8704, 2020.

\bibitem[Wang et~al.(2023)Wang, Bao, Zhou, Wang, Hu, Chen, and
  Li]{wang2023dire}
Zhendong Wang, Jianmin Bao, Wengang Zhou, Weilun Wang, Hezhen Hu, Hong Chen,
  and Houqiang Li.
\newblock Dire for diffusion-generated image detection.
\newblock In \emph{CVPR}, pages 22445--22455, 2023.

\bibitem[Yu et~al.(2022)Yu, Ni, Li, and Zhao]{yu2022detection}
Yang Yu, Rongrong Ni, Wenjie Li, and Yao Zhao.
\newblock Detection of ai-manipulated fake faces via mining generalized
  features.
\newblock \emph{ACM TOMM}, 18\penalty0 (4):\penalty0 1--23, 2022.

\bibitem[Zhang et~al.(2019)Zhang, Karaman, and Chang]{zhang2019detectingSPEC}
Xu Zhang, Svebor Karaman, and Shih-Fu Chang.
\newblock Detecting and simulating artifacts in gan fake images.
\newblock In \emph{WIFS}, pages 1--6. IEEE, 2019.

\bibitem[Zhong et~al.(2023)Zhong, Xu, Qian, and Zhang]{PatchCraft}
Nan Zhong, Yiran Xu, Zhenxing Qian, and Xinpeng Zhang.
\newblock Patchcraft: Exploring texture patch for efficient ai-generated image
  detection.
\newblock \emph{arXiv preprint arXiv:2311.12397}, 2023.

\bibitem[Zhu et~al.(2023)Zhu, Chen, Yan, Huang, Lin, Li, Tu, Hu, Hu, and
  Wang]{zhu2023genimage}
Mingjian Zhu, Hanting Chen, Qiangyu Yan, Xudong Huang, Guanyu Lin, Wei Li,
  Zhijun Tu, Hailin Hu, Jie Hu, and Yunhe Wang.
\newblock Genimage: A million-scale benchmark for detecting ai-generated image.
\newblock \emph{NeurIPS}, 2023.

\end{thebibliography}
